\documentclass[twoside]{article}

\usepackage[accepted]{mock}
\usepackage{times}
\usepackage{epsfig}
\usepackage{fancyhdr,graphicx,amsmath,amssymb}
\usepackage{xcolor,xspace}
\usepackage[round]{natbib}
\usepackage{algorithm}
\usepackage[noend]{algpseudocode}
\usepackage{amsmath}
\usepackage{amssymb}
\usepackage{booktabs}
\usepackage{balance}

\usepackage{amsfonts}      
\usepackage{url}

\newcommand{\comment}[1]{}

\newcommand{\ie}{\emph{i.e.},~}

\begin{document}

% If your paper is accepted and the title of your paper is very long,
% the style will print as headings an error message. Use the following
% command to supply a shorter title of your paper so that it can be
% used as headings.
%
%\runningtitle{I use this title instead because the last one was very long}

% If your paper is accepted and the number of authors is large, the
% style will print as headings an error message. Use the following command to supply a shorter version of the authors names so that
% they can be used as headings (for example, use only the surnames)
%
%\runningauthor{Surname 1, Surname 2, Surname 3, ...., Surname n}

\twocolumn[

\aistatstitle{Adaptive and Iteratively Improving Recurrent Lateral Connections}

\aistatsauthor{ Barak Battash \And Lior Wolf }

\aistatsaddress{ Tel Aviv University \And  Facebook AI Research \& Tel Aviv University } ]

\begin{abstract}
The current leading computer vision models are typically feed forward neural models, in which the output of one computational block is passed to the next one sequentially. This is in sharp contrast to the organization of the primate visual cortex, in which feedback and lateral connections are abundant. In this work, we propose a computational model for the role of lateral connections in a given block, in which the weights of the block vary dynamically as a function of its activations, and the input from the upstream blocks is iteratively reintroduced. We demonstrate how this novel architectural modification can lead to sizable gains in performance, when applied to visual action recognition without pretraining and that it outperforms the literature architectures with recurrent feedback processing on ImageNet.
\end{abstract}

\section{Introduction}

Rapid exposure experiments in primates teach us that image recognition  occurs as early as the first 100 msec of visual perception, a time budget that suffices only for feedforward inference, due to the relatively slow nature of biological neurons \citep{Perrett,Thorpe}. Feed forward models, which typically employ various skip connections, also seem to dominate the computer vision field of object recognition~\citep{krizhevsky2012imagenet,He2016DeepRL} and excel at tasks that were previously thought to require feedback, such as object detection and instance segmentation~\citep{He2017MaskR}, image segmentation~\citep{long2015fully,cciccek20163d}, object relation inference~\citep{dai2017detecting}, pose estimation of complex articulated objects~\citep{alp2018densepose}, and action recognition in video~\citep{Carreira_2017}.

However, anatomical studies have shown that feedback connections are prevalent in the cortex \citep{douglasmartin2004,fellemanvanessen1991}. As one striking example, the feedforward input from LGN to V1 in cats constitutes only five percent of the total input to V1, the rest being lateral and feedback connections \citep{Binzegger04}. In fact, lateral connections, which are projections from a layer to itself, are even more prevalent than feedback connections that project from downstream layers upstream.

One possible conjecture would be that feedback (including lateral) connections play roles that are replaced by other mechanisms in the current deep learning literature. For example, they could play a role in training the biological neural network, or they can form attention mechanisms, which are captured by attention~\citep{sermanet2014attention} and self-attention~\citep{parikh2016decomposable} blocks in modern neural networks. Similarly, one can claim that such connections are required due to the limitations of the biological computational, but may not be necessary in artificial neural networks, which can be extremely deep~\citep{liao2016bridging}. In all these cases, developing computational models for feedback connections would not necessarily impact machine learning (but would benefit brain science).

An alternative view would be to consider the power of lateral and feedback connections as underutilized mechanisms, for which the current literature does not provide practical and useful models. If this hypothesis is correct, the development of computational feedback models would lead to better treatment of tasks, in which the current computational models are still inaccurate and supply tools for improving the performance of the current models.

In this work, we consider lateral feedback connections as a mechanism for adapting the computation based on the specific input (i.e., the sample $x$). This is done in an iterative manner, where at each iteration $i$, the input from the upstream computations $u=e(x)$, $e$ being the upstream network, is passed through the recurrent block $f$ which has weights $W_i$. The activations that are produced $f(u,W_i)$ lead to new weights of this block, which are computed by a function $h$ as $W_{i+1}=h(f(u,W_i))$. The next time the same input $u$ passes through the layer, the layer is modified and produces a new set of activations $f(u,W_{i+1})$.

This novel type of recursion can be seen as a ``self-hypernetwork''.  Hypernetworks are neural networks in which the weights of some of the layers vary dynamically, based on the input. Such networks have composite architectures in which one network predicts the weights of another network. 
In our work, we extend this idea in few important ways. First, instead of separating into two networks, we add a second head ($h$) to the block, which determines the modified weights of the primary part of the block ($f$). Second, we employ the method iteratively multiple times, making the framework dynamic not just with respect to the varying sample $x$, but also vary from one iteration $i$ to the next, each time step, the recurrent block ``reflects'' on the same input $u$.

In addition, we encourage the network to become more accurate with each iteration, by introducing what we call the kaizen loss. Kaizen, which literally means improvement in Japanese, is often used to describe an ongoing process that leads to gradual and continuous positive changes. The kaizen loss we introduce  requires that the accuracy after iteration $i+1$ is better than the accuracy after iteration $i$. To avoid this loss being minimized by reducing the accuracy of the previous iterations, losses are put in place to maximize the accuracy after each iteration.

We focus on lateral connections in the last convolutional blocks of the network. While the architecture we propose is general, this focus allows us to maintain a reasonable number of experiments. In other words, since the modification of multiple blocks and forming feedback connections between blocks and other, upstream, blocks would call for multiplying the number of experiments by one or more factors, these are left for future work.

Our experiments are done in two domains: image recognition on benchmark datasets including MNIST and ImageNet and action recognition in video. While our method is general and can be applied in any domain, we focus on visual processing, since the visual cortex is well mapped and known to contain lateral connections. Video action recognition was chosen, since temporal processing is long thought to necessitate feedback and since it requires prolonged exposure due to the nature of the input. In addition, machine performance in the field of video action recognition is much lower than the human counterpart and the best current artificial systems either employ gigantic training datasets or combine multiple methods, only some of which rely on deep learning.

To summarize, our contributions are: (i) a novel self-hypernetwork block, which employs a single network with two heads, (ii) a computational model for lateral feedback connections that is based on an adaptive recurrent computation, (iii) the introduction of a new loss that leads to improvement from one iteration to the next, (iv) state of the art performance in two action recognition benchmarks in the case of no pretraining on outside data. (v) state of the art performance on ImageNet among similarly sized ImageNet architectures, and among the literature architectures with recurrent feedback connections.

\section{Related Work}

\paragraph{Earlier computational models for the role of feedback}

While the role of feedback in the brain is yet unknown, this question has inspired multiple computational approaches. In the biological literature, there is an emphasis on attention that is mediated by feedback~\citep{Desimone:ARN:95,decolee2002,vanRullen:JCN:01}. This view is congruent with cognitive experiments that show that when searching for a red ball, an emphasis is put on circular shapes and the red color. In such a case, recognition, given the attention, is performed by a feed-forward mechanism. As noted by \cite{leemumford2003}, this view does not capture all aspects of attention.

A related hypothesis is that attention plays a role in combining  bottom-up and top-down processing. \cite{epshteinullman2005} combine image patches in a bottom-up process to create a hierarchy of object parts of increasing complexity. At the top of the hierarchy, the object is recognized, and the information is propagated back to provide an exact localization.  \cite{borenstein2004} perform semantic segmentation by combining bottom up segmentation with top-down class-specific information. 

A specific form of combining bottom up and top down information that is attributed to feedback is the hypothesis verification scheme~\citep{carpentergrossberg1987,Mumford:BC:92,Hawkins04}. \cite{rosales2001,curiogiese2005} generate multiple pose hypothesis in a bottom-up pipeline, and then verify these by synthesizing a candidate image for each hypothesis. Such synthesis can be performed by differential renderers that are either geometric, such as the one introduced by  \cite{smelyansky2002dramatic} for optical flow computation, or, more recently, by~\cite{kato2018neural,meshrcnn} for 3D mesh reconstruction from a single image. They can also be learned, such as the face model of~\cite{blanz02}.% or the parameterized cartoon model of~\cite{wolf2017unsupervised}.

Another perspective on the role of feedback arises in probabilistic generative models, which explicitly model the probability $P(x|y)$ of an input image $x$ given an interpretation $y$. \cite{leemumford2003} employ an undirected graphical model to model the joint probability of the layers of the visual cortex. In their model, following \cite{sudderth2002}, the messages passed by the belief propagation method from a layer to the previous layer, can be seen as a form of feedback. 

\paragraph{Recurrent models inspired by feedback connections}

\cite{liao2016bridging} have studied the equivalence between RNNs and residual networks (ResNets)~\citep{He2016DeepRL} with shared weights. It is hypothesized that RNNs are biologically plausible mechanisms that can benefit from the performance boost obtained by deep ResNets. 

\cite{zamir2017feedback} incorporate RNN layers into their networks, which, like our method, receive the visual input repeatedly. Unlike our method, the layer does not change between iterations, is not conditioned on the input, and no gradual improvement in accuracy is enforced. However, the notion of taxonomic
prediction is introduced, in which subsequent iterations are optimized for distinguishing between classes at a finer scale, i.e., further down the taxonomy tree.

\cite{nayebi2018task} learn recurrent layers that capture both lateral connections (local recurrent feedback) as well as backprojections (long ranged feedback). Evaluation is done on ImageNet~\citep{imagenet} and two models were presented. The first model is a recurrent architecture, which employs gating (similar to LSTMs) and skip connections (similar to ResNets). This model performed similar to feed-forward networks that have 20\% more parameters and once optimized (thousands of hyperparameter experiments) the number of parameters dropped slightly and performance improved by a few percent. The second model was obtained by performing a large scale architecture search of the recurrent units. This model was shown to almost match the performance of ResNet-34, while having only $75\%$ of the parameters. While the gain in performance obtained is impressive and done on a large scale benchmark, these improvements are obtained  after performing an extremely large-scale hyperparameters or model search. Our results are obtained using a novel type of a recurrent architecture, which is applied across datasets and tasks.%and different baseline architectures. %On the competitive field of action recognition without using outside training data, our method obtains state of the art results.

\cite{leroux2018iamnn} presented slightly inferior results on ImageNet, in comparison to~\cite{nayebi2018task}, while using a third of the parameters and a relatively simple hand-engineered architecture that does not employ gating. In addition to the use of an RNN within the residual block, an early stopping mechanism is used. This mechanism is coupled with a dedicated loss term that encourages fewer iterations and the typical number of iterations is considerably lower than the maximal permitted one.

\cite{cichy2019deep} have presented both imaging evidence and a computational model to support the utility of recurrent processing for recognition under occlusion. Their hierarchical recurrent ResNet model outperforms AlexNet~\citep{krizhevsky2012imagenet} when recognizing heavily occluded objects.

\cite{cornet} develop a relatively shallow neural network with recurrent feedback connections, whose design is guided by a score that combines the correlation obtained between the network results and various biological neural and behavioral benchmarks. The correlation score obtained is higher than any other literature architecture and the recognition rates are high relative to the number of layers used. %However, in the number of parameters in their architecture is relatively high.

\paragraph{Video action recognition}

Previous work on feedback-inspired recurrent connections was able to show some improvement in comparison to baseline models without recursion and in comparison to feedforward models with a similar number of parameters. However, humans excel in image recognition even at short exposure times, \ie without employing feedback. In this work, we attempt to work in a domain in which the state of the art accuracy can be improved upon by incorporating feedback, regardless of the number of parameters, the task cannot be done effectively in rapid exposure settings, and humans greatly outperform machines. The domain of human action recognition in video presents these desiderata.

Deep learning models for action recognition can benefit from both image data and motion data~\citep{jhuang2007biologically,simonyan2014two}. 3D convolutions, which can capture temporal and spatial information, are prevalent in this domain~\citep{Ji_2013,Tran_2015}. Lagging far behind human level recognition, especially in comparison to image recognition, researchers have turned into using larger and larger datasets~\citep{abuelhaija2016youtube8m,Carreira_2017,kay2017kinetics,Monfort_2019}.

Meanwhile, other researchers have been focusing on learning with a limited amount of supervision, training from scratch on the UCF101 and HMDB51 datasets.
UCF101~\citep{soomro2012ucf101} has 13,000 videos for training and validation, whereas HMDB51~\citep{Kuehne_2011} is even smaller and contains 7,000 videos for training and validation. Similarly to most other recent work, we use the first split of those datasets in our experiments.

\cite{2017arXiv171109577H} demonstrated that training a Conv3D network on these small video datasets leads to overfitting. \cite{2019arXiv190704632P} use network architecture search in order to find suitable architectures.  \cite{2018arXiv181111387J} use the very large kinetics~\citep{kay2017kinetics} or moments in time~\citep{Monfort_2019} datasets to pretrain their model in a self-supervised manner by predicting the set of rotations applied to a video. Following this pretraining they finetuned their model on HMDB51 and UCF101. \cite{2018arXiv180607754D} presented a spatio-temporal channel correlation block that extracts the relations between different channels in the intermediate feature maps.

\paragraph{Hypernetworks} 

The idea of dynamic layers, whose weights vary based on the input and are determined by a second neural network, first appeared as a way to adapt the lower layers to the motion or illumination of the image input \citep{klein2015dynamic,7410424}. A wider application of the idea by \cite{jia2016dynamic}, employs hypernetworks across multiple layers, for video frame synthesis and stereo views prediction.

RNNs in which the weights are determined by another RNN were presented by~\cite{ha2016hypernetworks}. In this case, the weight generating RNN receives both the previous hidden state and the next token as its input. Unlike our method, the two networks are disjointed and their input vary over time. The Bayesian formulation of \cite{krueger2017bayes} introduces variational inference that involves a parameter generating network and a primary network. 

\cite{bertinetto2016learning} use hypernetworks for few-shot learning tasks, utilizing the ability of the network to adapt to the current task and the ability to share knowledge between different tasks via the weight generating network. The ability of hypernetworks to replace backpropagation-based training by prediction of weights was exploited by \cite{brock2018smash,zhang2018graph} for performing architecture search.

\begin{figure}[t]
  \centering
  \begin{tabular}{cc}
  \includegraphics[width=0.40\linewidth]{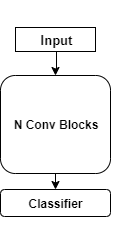}&
  \includegraphics[width=0.553\linewidth]{Images/HighLevelDyRect.png}\\
  (a) & (b)\\
  \end{tabular}
  \caption{The generic way to modify an existing neural network. (a) The original feed forward network with $N$ convolutional blocks, each containing one or more layers. (b) The application of our method: the last $R$ convolutional blocks are augmented with recurrent connections. At each recurrent iteration, the output of block $N-R$ is passed to the first recurrent block. Once the $K$ iterations of the first recurrent block are completed, this block's output is passed to the second recurrent block and so on.}
  \label{fig:highlevel}
\end{figure}

\section{Method}

Given a baseline architecture of a feed-forward network with $N$ convolutional blocks, as depicted in Fig.~\ref{fig:highlevel}(a), we modify the last $R$  blocks, by adding recurrent connections to it, see Fig.~\ref{fig:highlevel}(b). A convolutional block would typically contain multiple convolution layers and residual connections. We focus on the last blocks in order to emphasize high-level processing and to limit the scope of the paper as the means of producing a manageable number of experiments.

The modified, recurrent, blocks consist of three main components: the primary subblock $f$, the optional static subblock $g$ (which can be the identity), and the weight modifying subblock $h$. Each of these computational components is a function of a single (tensor) input. However, the weights of $f$ vary between iterations and we, therefore, write it as $f(u,W_i)$, where $u$ is the input to the recurrent block (the result of the upstream computation) and $W_i$ are the weights of this block at iteration $i$.

\begin{figure}[t]
  \centering
  \includegraphics[trim=75 0 40 167, clip,width=0.81595\linewidth]{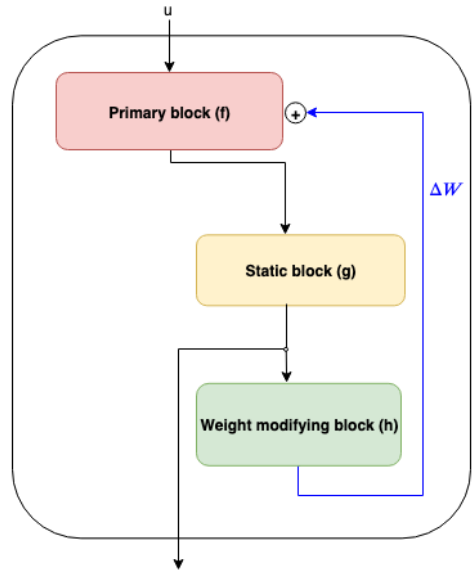}
  \caption{An illustration of an adaptive recurrent block. The black arrows represent the data flow; the blue arrows represent the weights dynamics. The input $u$ to the block is unchanged during the recursions and is the result of upstream computation. The primary block $f$ employs the dynamic weights $W_i$ to perform a computation given this input. The optional static block $g$ is a fixed-weights network that is downstream from the primary block. The weight-adaptation block $h$ is downstream from the static block, and it computes the change of the weights $\Delta W$ in the primary block of the next iteration.}
  \label{fig:generalblock}
\end{figure}

Fig.~\ref{fig:generalblock} illustrates the structure of the recurrent block. The subblocks $f$ and $g$ are obtained by dividing the layers of the original non-recurrent block into two parts: a dynamic part, in which the weights vary as a function of the input and between iterations, and a static one.

Let $z_i$ be the output of the recurrent block at iteration $i$. It is given by 
\begin{equation}
    z_i = g(f(u,W_i))
\end{equation}
where $W_1$ is the list of initial weights of the neural network $f$, and $W_i$ for $i>1$ are described below. Note that $g$ (and $h$) also have learnable weights, but we do not write these explicitly, since the weights do not change dynamically.

The weights of the primary block vary between iterations according to the recursion formula
\begin{equation}
    W_{i+1} = W_i + \alpha h(z_i).
    \label{eq:wi}
\end{equation}
\ie the weights are updated additively by the output of network $h$ multiplied by a hyperparameter $\alpha>0$. $W_i$ for $i>1$ are conditioned on $u$ through the activations $z_i$. %factor that is $\alpha>0$. % at the first update and is then decreased by a factor of $\beta \in [0,1]$, where $\alpha$ and $\beta$ are hyperparameters. %{\color{red} Not Learned}

Note that throughout all iterations, the input $u$ to the recurrent block does not change. In the case in which $R>1$ the input of the $r+1$ recurrent block is the output of the $r$th recurrent block after all $K$ iterations are completed, for $r=1,2,\dots,R-1$.

\subsection{Training}

The training takes place in two phases. In the first phase, we train the unmodified network that contains: (i) the first $N-R$ convolutional blocks, which we refer to as the network $e$, (ii) the last $R$ non-recurrent convolutional blocks, which contain the layers that would become $f$ and $g$, and (iii) the downstream classifier $c$. 

For simplicity, assume that in the feedback-augmented network there is one recurrent block ($R=1$), which is the $N$th block. In the case of additional recurrent blocks, each one is trained with its own weights.  

Similar to the notation in the network with the recurrent block, we denote the computation that is being performed by the $N$th block of the first phase on input $u$ as $g(f(u,W_1))$. This notation indicates the division, when the recurrent block is introduced in the second phase, of the $N$th convolutional block into two parts that correspond to the primary block $f$ and the static block $g$. The entire network computation in the first phase, on some input $x$, is thus given by $c(g(f(e(x),W_1)))$.

In the second phase, $g$ is trained with backpropogation, and the weights of $f$ change dynamically, using a new network block $h$.  The initial weights of $f$, called $W_1$, remain fixed after the first phase of training. The weights of $g$ are initialized by the corresponding values in the $N$th convolutional block, as obtained at the end of the first phase, and continue to train in the second phase. $h$ is initialized at random.

To be clear: the sets of weights $W_i$ are not directly trained during the second phase, and are obtained, for $i>1$, from $h$. These weights are a function of $W_1$ that is fixed from the first phase, and change (for $i>1$) from one sample $x$ to the next, due to the dependence of $h$ on the input of $f$.

Recall that $c$ is the classifier that is applied after the $N$th convolutional block. We assume that $c$ outputs a vector of pseudo-probabilities, one per class. Recall also that $e$ denotes the computation that takes place at the first $N-1$ convolutional blocks, \ie $u=e(x)$ for the input sample $x$. In the first phase of training, the cross entropy loss is used:
\begin{equation}
    \mathcal{L}^I = - \sum_j \log c_{y_j}(g(f(e(x_j),W_1)))
\end{equation}
where the training samples and the corresponding labels are given as $(x_j,y_j)$, the index $j$ runs over all training samples, and the subscript of $c$ denotes a single element of this classifier's output.

In the second phase, the cross entropy loss is applied repeatedly, once per each iteration $i=1..K$. 
\begin{equation}
    \mathcal{L}_i = - \sum_j \log c_{y_j}(g(f(e(x_j),W_i)))
    \label{eq:iterlosses}
\end{equation}

The loss of phase two is the sum of all these losses with an additional loss that encourages the loss to decrease from one iteration to the next. The overall loss is given by
\begin{equation}
    \mathcal{L}^{II} = \mathcal{L}_K +  \sum_{i=1}^{K-1} \gamma_i \mathcal{L}_i + \lambda \sum_{i=1}^{K-1}  \max(0,\mathcal{L}_{i+1}-\mathcal{L}_{i}) 
    \label{eq:phasetwoloss}
\end{equation}
where $\gamma_i \in [0,1]$ are parameters that are added to reflect that the final loss $\mathcal{L}_K$ is more important that the other cross entropy losses. The third term of the loss is weighted by another parameter $\lambda>0$ and is called the kaizen loss. It adds a penalty, if the loss of the previous iteration is lower than the loss of the current iteration. 

The loss $\mathcal{L}^{II}$ is minimized with respect to the weights of $e$, $g$, $h$, and $c$, where the weights of $h$ are manifested in the loss through the weights $W_i$ of $f$, as described in Eq.~\ref{eq:wi}.

Note that in the case of $R>1$, the loss is applied only to the last block (the $N$th block), which receives the output of block $N-1$ after the $K$ iterations are completed (also true for all upstream recurrent blocks). Therefore, no loss enforces improvement from one iteration to the next for the blocks before the $N$th block. However, all are required to perform well after $K$ iterations.

% \subsection{Dropout}
% Like most CNNs, MFNet \cite{Chen2018} is also using dropout layer before the classifier in order to prevent overfitting.
% While using Feedback mechanism it is hardening (?) from the network to train. in order to solve this issue we decreased the dropout to a lower value, By using the next thumb rule: more loops less dropout.
% This path helped us to make a better training but now the but now the overfitting is around the corner.
% We splitted the dropout, while using a low value dropout layer inside the \feedback loop, we also used a dropout layer with a higher value before the entering the \feedback loop, this helped crucially to obtain higher results.

\section{Experiments}

We conduct experiments on image recognition as well as on action recognition in video. First, we employ the MNIST benchmark and focus on comparing with the baseline architectures. In action recognition, we achieve state of the art results on both HMDB51 and UCF101. Finally, in order to compare with recent work that employ recurrent feedback mechanisms, we provide ImageNet experiments.

In all of our experiments $R=1$ and $\alpha=10^{-3}$. We set $\gamma_1=3\times 10^{-3}$ and $\gamma_i= 10^{-3}$ for $i=2,3,..,K-1$, since we found it to be beneficial to emphasize the loss of the first iteration as well as the last one (recall from Eq.~\ref{eq:phasetwoloss} that the cross entropy loss of iteration $K$ is taken without a factor). The parameter of the kaizen loss is taken as $\lambda \sim 0.1$, except in our ablation experiments, where it is set to $0$.

\subsection{Vanilla CNN experiments on MNIST}

For the MNIST benchmark by \cite{lecun-mnisthandwrittendigit-2010}, we employ a simple CNN with two convolution layers and two fully connected layers and ReLU activations. Both conventional layers employ kernels of size $5\times 5$. The first convolutional layer has one input channel, and three output channels. The second also has three output channels. 

We consider each layer (convolution or fully connected, which is a specific case of a convolution layer) to be a block by itself, and thus the network has $N=3$, and a one layer classifier $c$. We experiment only with $R=1$, i.e., turn the first fully connected layer into a recurrent layer. This layer has 48 input neurons and 10 output neurons. We take $f$ to be the entire third layer, and $g$ to be the identity. $h$ has a single layer of 10 input neurons and 480 output neurons. The second fully connected layer, which plays the role of $c$, has $10 \times 10$ parameters. No bias terms are used. 

In addition to our recurrent model, for 2-4 iterations, we also report our results without the kaizen loss (i.e., $\lambda=0$ in Eq.~\ref{eq:phasetwoloss}). Two baselines are also reported: one is the model performance after phase one, and one is an enlarged model in which the number of parameters matches the one of our recurrent model. The latter baseline, which we term ``baseline-big'' in our results, has five 2D convolution layers, followed by two fully connected layers, for a total of 5687 neurons.

The results are given in Tab.~\ref{tab:mnist}. As can be seen, the additional recurrent layers improve performance over the baseline method and the results gradually improve with the iterations. The kaizen loss seems to contribute to the accuracy. The gap in performance in comparison to the parameter-matched (``big'') baseline is lower. However, our full method outperforms it as soon as the first recurrent iteration occurs. 

\begin{table*}[t]
  \caption{Results on the MNIST dataset. No recurrent iterations (-) implies one iteration through each block.}
  \label{tab:mnist}
  \centering
\begin{tabular}{lccc}
\toprule
Method          & Recurrent iterations  &  Top-1 accuracy  & Number of parameters \\
\midrule
Baseline (phase one of our method) & -  &  96.50                   &  900                   \\
Baseline-big   & -&  98.07                   &  5687                   \\
\midrule
Our recurrent connections, no kaizen loss & 2  &        98.03             &  5691                   \\
Our recurrent connections, no kaizen loss & 3  &        98.15            &  5691                   \\
Our recurrent connections, no kaizen loss & 4  &        98.16         &  5691                   \\
\midrule
Our full method & 2  &        98.16             &  5691                   \\
Our full method & 3  &        98.32            &  5691                   \\
Our full method & 4  &        98.43            &  5691                   \\
\bottomrule
\end{tabular}
\end{table*}

\subsection{MFNet experiments on HMDB51 and UCF101}
\label{sec:expaction}
\begin{figure}[t]
  \centering
  \includegraphics[trim=0 17 80 17, clip,width=0.82342479\linewidth]{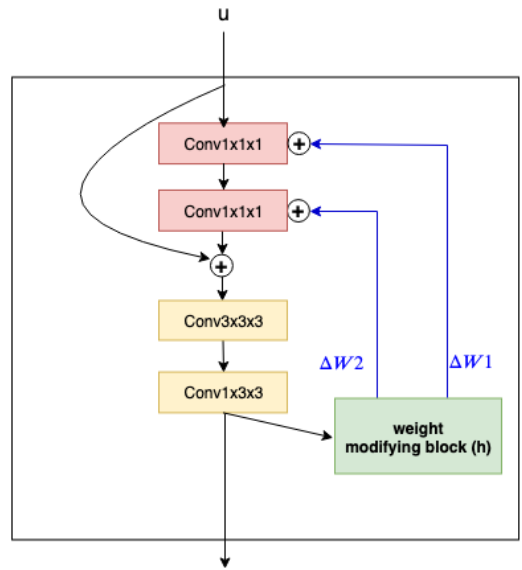}
  \caption{A MultiFiberNet block with our recurrent mechanism. $f$ consists of the first two convolution layers.}
  \label{fig:mfnetblock}
\end{figure}

\begin{table*}[t]
  \caption{Results on the UCF101 dataset. The literature results are obtained from \citep{2018arXiv180700230K} }
  \label{tab:ucf}
  \centering
\begin{tabular}{lcccc}
\toprule
Method & Iterations  &  Clip acc. & Top-1 acc. & Num. parameters [M]\\
\midrule
NAS \citep{2019arXiv190704632P} & -  & & 58.6 &  0.67\\
STC-ResNet 101 \citep{2018arXiv180607754D} &- &  & 47.9 & 100+ \\
Long-term Temporal Convolutions 60f \citep{2016arXiv160404494V}  &-& 57.0& 59.9& \\
3D-ResNet18 \citep{2017arXiv171109577H} &-  &  &42.4 &33.2 \\ 
Video Rotation Prediction model (kinetics) \citep{2018arXiv181111387J} &- & &62.9&11.7\\
Video Rotation Prediction model (MT) \citep{2018arXiv181111387J} &-& &62.8&11.7\\
Res3D \citep{10.1007/978-3-030-01264-9_39} &-&&67.6& 33.2+\\
\midrule
Baseline (phase one of our method)  & -  & 64.3      &  67.4          &  0.5                   \\
Baseline-big     & -&  68.0  & 72.5      &  7.7                   \\
\midrule
Our recurrent connections, no kaizen loss& 2 & 66.3 & 73.1                    &  0.63                   \\

Our recurrent connections, no kaizen loss & 3 & 67.8 & 73.9                     & 0.63                     \\

\hline
Our full method  & 2 & 67.3 &  74.3                   &  0.63                   \\

Our full method & 3 & 68.3 &  73.6                    &  0.63                   \\

\bottomrule
\end{tabular}
\end{table*}
\begin{table*}[t]
  \caption{Results on the HMDB51 dataset. The literature results are copied from \citep{2018arXiv180700230K}}
  \label{tab:hmdb}
  \centering
\begin{tabular}{lcccc}
\toprule
Method          & Iterations  & Clip acc. &  Video acc. & Num. parameters [M] \\
\midrule
ResNet18-3D\citep{2017arXiv171109577H} &- && 17.1 &33.2 \\ 
Video Rotation Prediction model (kinetics) \citep{2018arXiv181111387J} &-&&33.7&11.7\\
Video Rotation Prediction model (MT) \citep{2018arXiv181111387J} &-&&29.6&11.7\\
Res3D \citep{10.1007/978-3-030-01264-9_39} &-&&33.1& 33.2+\\
\midrule
Baseline (phase one of our method) & -  & 28.8  &    31.9                 &  0.5                   \\
Baseline-big    & - & 28.8 & 31.3   &        7.6             \\
\midrule
Our recurrent connections, no kaizen loss & 2  & 33.9  & 37.3                   &  0.62                   \\
Our recurrent connections, no kaizen loss & 3  & 34.2 & 38.4                   &  0.62                   \\
%Our recurrent connections, no kaizen loss & 4 & &                     & 0.62                     \\

\midrule
Our full method & 2  & 33.9 & 37.6                   &  0.62                   \\
Our full method & 3  & 34.7 &38.8                 &  0.62                   \\
\bottomrule
\end{tabular}
%\end{table*}
%\begin{table*}[t]
\medskip 
%\smallskip
%\vspace{.00251cm}
  \caption{Results on the ImageNet dataset.}
  \label{tab:imagenet}
  \centering
\begin{tabular}{lcc}
\toprule
Method             & Number of parameters [M] &  Top-1/Top-5 accuracy\\
\midrule
\multicolumn{3}{c}{-- No feedback literature baselines --}\\
ResNet152~\citep{He2016DeepRL}  	 & 60   & 77.0 / 93.3	\\
ShaResNet34~\citep{boulch2018sharesnet}  & 14			& 71.0 / 91.5 	\\
ResNet18~\citep{He2016DeepRL}	 & 12   	   	& 69.5 / 89.2  \\
Googlenet~\citep{szegedy2015going}     & 7	 		& 65.8 / 87.1 	\\
ShuffleNet2x~\citep{zhang2018shufflenet}  & 5.6  		& 70.9 / 89.8  	\\
MobileNet1~\citep{howard2017mobilenets}    & 4.2  		& 70.6 / 89.5 	\\
SqueezeNet~\citep{iandola2016squeezenet}    & 1.3    &  57.5 / 80.3	\\
\midrule
\multicolumn{3}{c}{-- Literature baselines with recurrent feedback architectures --}\\
ConvRNN~\citep{nayebi2018task} &  15.5 & 72.9/ unreported \\
IamNN 1 iter~\citep{leroux2018iamnn} & 4.8 		& 60.8 / 83.2  \\
IamNN~\citep{leroux2018iamnn}      	 & 5 			& 69.5 / 89.0	\\
CORnet-S, most correlated with brain activity~\citep{cornet} & 53.4 & 73.9 / unreported\\
CORnet-S, best epoch~\citep{cornet} & 53.4 & 74.4 / unreported \\
\midrule
Baseline (phase one of our method, i.e., MFNet2D) &     5.72  & 73.9/91.8 \\
Our full method (2 iterations)  &    5.91  &   74.5/92.0      \\
Our full method (3 iterations)  &    5.91  &   74.7/92.1      \\
%added by barak after submission
Our full method (4 iterations)  &    5.91  &   74.8/92.1      \\

\bottomrule
\end{tabular}
\end{table*}    

Action recognition evaluation is often reported at the video level. However, training is done at the clip level, each clip consisting of a fixed number of frames. In our experiments, video level prediction is obtained by randomly sampling ten clips from each video, each clip consisting of 16 frames. The probabilities from each clip are summed to perform a voting at the video level.

Our action recognition experiments employ the MFNet3D architecture by \cite{Chen2018}. The architecture consists of five high level blocks, each containing between one and three MultiFiberNet blocks. In order to reduce memory consumption, we employ a version with a total of $N=9$ MultiFiberNet blocks. When employing our method, we convert the last one into a recurrent block ($R=1$) as described below.

A typical MultiFiberNet block consists of four convolution layers: two 3D convolution layers with a kernel of size $1\times 1 \times 1$ (can be considered as dimensionality reduction), followed by two  group 3D convolution layers, where the channels are split into eight groups, each one processed independently, and the results are concatenated. The first group convolution has kernels of size $3\times 3\times 3$, and the second has $1\times 3\times 3$ kernels, where the first dimension is the temporal one. 

We consider the function $f$ that consists of the two $1\times 1 \times 1$ layers, and a function $g$ that consists of the group convolutions, as depicted in Fig.~\ref{fig:mfnetblock}. 

The weight modifying function $h$ consists of a max pooling layer with a kernel of size $5\times 3\times 3$ with a stride of (4,1,1). In parallel, an average pooling with the same kernel size is used. The outputs of two layers are concatenated and vectorized and are followed by one fully connected layer for predicting the increment of the weights of the first dynamic layer, and one for updating the second. 

%In order to avoid predicting a larger number of parameters, we apply the same shift to each $1\times 1\times 1$ kernel, thus the output of the weight predicting function $h$ is a vector in $\mathbb{R}^48$ for updating the first layer of the MultiFiberNet block and a simular sized vector for updating the second layer.

As before, a first baseline is taken to be the network after the first phase of training (no recurrent iterations). The big baseline is taken to be our implementation of the original MFNet3D network. This baseline has an order of magnitude more parameters than our recurrent network. Note that unlike the results reported by \cite{Chen2018}, we do not employ pretraining on an outside dataset. 

The results for the UCF101 benchmark are reported in   Tab.~\ref{tab:ucf}. As can be seen, our method outperforms the other methods by a sizable gap, despite using much fewer parameters. Our method outperforms not only models trained from scratch but also the two models of \cite{2018arXiv181111387J},  which were pretrained on very large datasets in a self supervised manner. Furthermore, \cite{2016arXiv160404494V} use 60 frames as input, while we only use 16 frames. 

The results show that the first recurrent iteration adds considerably more accuracy than the second, which also has a positive contribution. The positive contribution of the kaizen loss is also evident. The one exception is the case of three iterations, where the kaizen loss reduces the Top-1 accuracy.  However, in this case as well, it seems to increase the clip accuracy, which is the score that the network is trained to optimize.

The results for the HMDB51 benchmark are reported in   Tab.~\ref{tab:hmdb}. As can be seen, our method outperform the previous methods by a large margin. As in the UCF101 dataset, the method of \cite{2018arXiv181111387J} is pretrained on the large kinetics or moments in time dataset in a self-supervised way. Nevertheless our model outperforms their best result by 5.1\% in the video level accuracy. There is also a noticeable gap between our baseline models and our full method. This gap is despite our model having only slightly more parameters than the baseline model and much fewer than the baseline-big model.

We also note that the bigger baseline is not more accurate than the smaller. This demonstrates the overfitting challenge on small datasets, and further emphasizes the success of our model in improving the baseline model by almost 7\%. This overfitting phenomenon was not observed on the somewhat bigger and much easier UCF101 dataset.

\subsection{MFNet experiments on ImageNet}
Finally, in order to compare with previous work that applied feedback powered networks on ImageNet, we employ the same MultiFiberNet architecture with $N=9$  used in the action recognition experiments, after replacing the 3D convolutions with 2D ones. As in all other experiments, we convert the last block to a recurrent block ($R=1$). The same recurrent block as in Fig.~\ref{fig:mfnetblock} is used,  with 2D convolutions in lieu of 3D ones.

The weight modifying block $h$ is similar to the one from Sec.~\ref{sec:expaction},  the two pooling layers (max and average) are applied with a kernel of size $3 \times 3$ and a stride of (1,1). The pooling results are concatenated and followed by two fully connected layers, one predicting the increment of the weights of the first dynamic layer, and one predicting the second. this configuration, $h$ consists of only 26k parameters.

The results are reported in Tab.~\ref{tab:imagenet}. As can be seen, our method outperforms all limited parameter networks and has a clear advantage in performance and the number of parameters in comparison to all literature baselines that employ recurrent feedback signals. Starting from the 2nd iteration, our method outperforms the most recent and accurate feedback-based method, despite having an order of magnitude less parameters. It is also evident that the performance improves from one iteration to the next.

\section{Conclusions}

The current state of the art deep learning methods do not incorporate self- and back-projections, which are prevalent in the visual cortex. In this work, we demonstrate that feedforward recognition algorithms can be substantially improved by adding lateral feedback recursions.

Our success suggests a possible hypothesis that addresses one of the major puzzles in visual sciences:  the role of feedback. We provide indirect evidence that the main role of lateral feedback is to allow the computation to adapt to the input stimulus, in a manner that improves accuracy. This view is an alternative view to that of using recursion for emulating deeper networks with shared weights (in our network the weights change but the same input $u$ is reintroduced), to the views that emphasize top-down and bottom-up integration (so far, we apply our method to the top layers and we focus on lateral connections),  to explicit hypothesis verification models, and to attention as it is currently understood (our design is very different from attention blocks).

{\small
\bibliographystyle{apalike}
\bibliography{hypernets,more,EMT,bb,contour}
}
\end{document}